\documentclass{article}

\usepackage{PRIMEarxiv}
\usepackage{hyperref}
\usepackage{float}
\usepackage{verbatim} 
\usepackage{graphicx}
\usepackage{xcolor}
\usepackage{booktabs}       
\usepackage{amsmath,amsthm,amssymb,amsfonts}
\usepackage{subcaption} 
\usepackage{listings}
\usepackage{lscape}
\usepackage{multirow}
\usepackage{bm}
\usepackage{dsfont}
\restylefloat{figure}
\restylefloat{table}

\newcommand{\E}{\mathbb{E}}
\newcommand{\bR}{\mathbb{R}}

\def\cD{\mathcal D}
\def\cF{\mathcal F}

\def\cL{\mathcal L}

\newcommand{\bX}{\mathbf{X}}

\newcommand{\bx}{\mathbf{x}}

\def\be{\begin{equation}}
	\def\ee{\end{equation}} 
\def\bi{\begin{itemize}}
	\def\ei{\end{itemize}} 
\def\ba{\begin{array}}
	\def\ea{\end{array}}

\theoremstyle{plain}%
\theoremstyle{definition}%

\title{Within-group fairness: A guidance for more sound between-group fairness}
\date{}
\author{
			Sara Kim\\
		Samsung Electronics Co.\\
		Seoul\\
		\texttt{sarah58833@gmail.com} \\
		\And
		Kyusang Yu \\
		Department of Applied Statistics \\
		Konkuk University \\
		Seoul\\
		\texttt{kyusangu@konkuk.ac.kr} \\
		\And
		Yongdai Kim \\
		Department of Statistics \\
		Seoul National University \\
		Seoul\\
		\texttt{ydkim0903@gmail.com} \\
}

\begin{document}
	\maketitle
	
	\begin{abstract}
		As they have a vital effect on social decision-making, AI algorithms not only should be accurate and but also should not pose unfairness against certain sensitive groups (e.g., non-white, women). 
		Various specially designed AI algorithms to ensure trained AI models to be fair between sensitive groups have been developed. 
		In this paper, we raise a new issue that between-group fair AI models could treat individuals in a same sensitive group unfairly.
		We introduce a new concept of fairness so-called {\it within-group fairness} which requires 
		that AI models should be fair for those in a same sensitive group as well as those in different sensitive groups. 
		We materialize the concept of within-group fairness by proposing corresponding mathematical definitions and developing learning algorithms to control within-group fairness and between-group fairness simultaneously. 
		Numerical studies show that the proposed learning algorithms improve within-group fairness without sacrificing accuracy as well as between-group fairness. 
	\end{abstract}
	

	\section{Introduction}\label{sec1}
	
	Recently, AI (Artificial Intelligence) is being used as decision-making tools in various domains such as credit scoring, criminal risk assessment, education of college admissions \cite{angwin2016machine}. As AI has a wide range of influences on human social life, issues of transparency and ethics of AI are emerging.  
	However, it is widely known that due to the existence of historical bias in data against ethics or regulatory frameworks for fairness, trained AI models based on such biased data
	could also impose bias or unfairness against a certain sensitive group (e.g., non-white, women) \cite{kleinberg2018algorithmic, mehrabi2019survey}. Therefore, designing an AI algorithm which is 
	accurate and fair simultaneously has become a crucial research topic.

	Demographic disparities due to AI, which refer to socially unacceptable bias that an AI model favors certain groups (e.g., white, men) over other groups (e.g., black, women), have been observed frequently in many applications of AI such as COMPAS recidivism risk assessment \cite{angwin2016machine}, Amazon's prime free same-day delivery \cite{ingold2016amazon}, 
	credit score evaluation \cite{Dua:2019} to name just a few.
	Many studies have been done recently to develop AI algorithms
	which remove or alleviate such demographic disparities in trained AI models
	so that they will treat sensitive groups as equally as possible.
	In general, these methods try to search AI models which are not only accurate but also
	similar between sensitive groups in a certain sense. For an example of similarity, it is required that
	accuracies of an AI model for each sensitive group are similar \cite{zafar2019fairness}.
	Hereinafter, criteria of fairness requiring similarity
	between sensitive groups are referred to as {\it between-groups fairness} (BGF).

	In this paper, we consider a new concept of fairness so called {\it within-group fairness} (WGF) which arises as a new problem when we try to enforce BGF into AI algorithms. 
	Generally speaking, within-group unfairness occurs when there is an individual who is positively treated compared to others in a same sensitive group by an AI model trained without BGF constraints but becomes negatively treated by an AI model trained with BGF constraints. 
	
\begin{figure}[t]
		\begin{center}
			\centerline{\includegraphics[width=0.55\columnwidth]{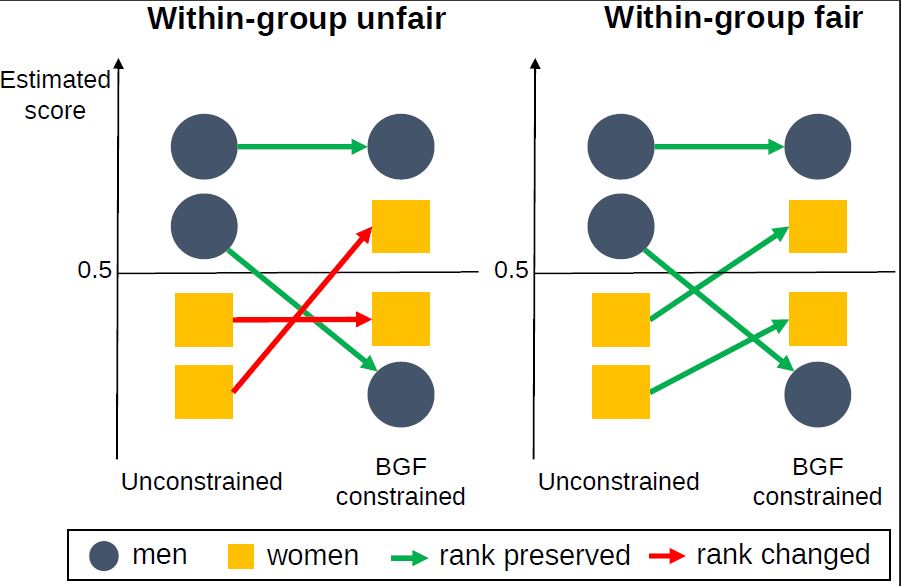}}
			\caption{A toy example of within-group unfairness: 
				The left panel: without BGF constraints, there exists unfairness against the women sensitive group, but with BGF constraints, the scores of the two women
				become reversed and thus within-group unfairness occurs.
				The right panel: The scores of the two women increase together to achieve BGF without
				within-group unfairness.}
			\label{fig:fig1}
		\end{center}
		\vskip -0.2in
	\end{figure}

	For an illustrative example of WGF, consider a college admission problem where gender (men vs women) is a sensitive variable. Let $\bX$ and $Y\in \{0,1\}$ be the input vector and the corresponding output label where $\bX$ represents the information of a candidate student such as GPA at high school, SAT score, etc. and $Y$ is the admission result where 0 and 1 mean the rejection and acceptance 
	of the college admission, respectively. The Bayes classifier accepts a student with $\bX=\bx$
	when $\Pr(Y=1\vert\bX=\bx)>1/2.$  Suppose that there are two women `$A$' and `$B$' 
	with the input vectors $\bx_A$ and $\bx_B,$ respectively and the AI model trained without BGF constraints
	estimates $\Pr(Y=1\vert\bX=\bx_B)>1/2 >\Pr(Y=1\vert\bX=\bx_A).$
	Then, within-group unfairness occurs when an AI model trained with BGF constraints
	results in $\Pr(Y=1\vert\bX=\bx_A)>1/2 >\Pr(Y=1\vert\bX=\bx_B).$  
	In this situation, which is illustrated in the left panel of Figure \ref{fig:fig1}, `$B$' could claim that the AI model trained with BGF constraints mistreats her and so it is unfair.  
	We will show in Section \ref{simulation} that there exists non-negligible within-group unfairness in AI models trained on real data with BGF constraints.

	Within-group unfairness arises because most existing learning algorithms for BGF force certain statistics (e.g. rate of positive prediction, misclassification error rate, etc.) of a trained AI model being similar across sensitive groups but do not care about what happens to individuals in a same sensitive group at all.
	For within-group fairness, a desirable AI model is expected at least to preserve the ranks between  
	$\Pr(Y=1\vert\bX=\bx_A)$ and $\Pr(Y=1\vert\bX=\bx_B)$ regardless of estimating $\Pr(Y=1\vert\bX=\bx)$
	with or without BGF constraints,
	which is depicted in the right panel of Figure \ref{fig:fig1}. 
	\medskip
	
	Our contributions are three folds. We first define the concept of WGF rigorously. Then we develop learning algorithms which compromise
	BGF and WGF as well as accuracy. 
	Finally, we show empirically that the proposed learning algorithms improve WGF while maintaining
	accuracy and BGF.
	\medskip
	
	\noindent{\bf Remark.} One may argue that training data are prone to bias
	due to historical prejudices and discriminations, and hence a trained AI 
	model is also biased and socially unacceptable. 
	On the other hand, a trained AI model with BGF constraints does not have such bias and
	hence is socially acceptable. Therefore, it would be by no means reasonable
	to claim unfairness based on discrepancies between socially unacceptable and acceptable AI models.
	However, note that historical bias in training data is about bias between sensitive groups but not for individuals in a same sensitive group. For WGF, we implicitly assume that
	no historical bias among individuals in a same sensitive group exists in training data, which is not too absurd,
	and thus there is no reason for a trained AI model without BGF constraints to treat individuals in a same sensitive group unfairly. This assumption, of course, needs more debates which we leave as future work.
	\medskip
	
	The paper is organized as follows. 
	In Section \ref{review}, we briefly review methods for BGF, and in Sections \ref{wgf-classifier} and \ref{wgf-score}, we propose
	mathematical definitions of WGF and develop corresponding learning algorithms
	for classifiers and score functions,  respectively.
	The results of numerical studies are presented in Section \ref{simulation}, and remarks about reflecting WGF to pre- and post processing algorithms for BGF are given in Section \ref{remarks}.
	Concluding remarks follow in Section \ref{conclusion}.

	\section{Review of between-group fairness}\label{review}

	While it is completely new, the concept of WGF is a by-product of BGF and thus it is helpful to
	review learning methods for BGF.
	In this section, we review the definitions of BGF
	and related studies.
	
	We let $\cD = \{(\bx_i, z_i, y_i)\}_{i=1}^n$ be a set of training data of size $n$ which are independent copies of a random vector
	$(\bX,Z,Y)$ defined on $\mathcal{X} \times \mathcal{Z} \times \mathcal{Y},$ where $\mathcal{X}\subset \bR^p.$
	We consider a binary classification problem, which means $\mathcal{Y}=\{0, 1\},$ and for notational simplicity, we
	let $\mathcal{Z}= \{ 0, 1 \}$, where $Z=0$ refers to the unprivileged group and $Z=1$ refers to the privileged group.
	Whenever the probability is mentioned, we mean it by either the probability of $(\bX,Z,Y)$ or its empirical counterpart unless there is any confusion.
	
	In this paper, we consider AI algorithms which yield a real-valued function $f:\mathcal{X}\rightarrow \bR$
	so called a score function which assigns positive labeled instances higher scores than negative labeled instances.
	An example of the score function is the conditional class probability $\Pr(Y=1\vert\bx=\bx).$
	In most human-related decision makings,  real-valued score functions are popularly used (e.g. scores for credit scoring).

	Let $\mathcal{F}$ be a given set of score functions, in which we search an optimal score function in a certain sense (e.g. minimizing the cross-entropy for classification problems).
	Examples of $\cF$ are linear functions, reproducing kernel Hilbert space and deep neural networks to name a few.
	For a given $f\in \cF,$ the corresponding classifier $C_f$ is defined as $C_f(\bx)=\mathds{1}(f(\bx)>0).$
	
	\subsection{Definition of between-group fairness}
	
	For a given score function $f$ and a sensitive group $Z = z$, we consider
	the group performance function of $f$ given as
	\be
	\label{eq:qz}
	q_z(f):=  \E(\mathcal{E}\vert\mathcal{E}', Z=z)
	\ee
	for  events $\mathcal{E}$ and $\mathcal{E}'$  that might depend on $f(\bX)$ and $Y.$ 
	The group performance function  $q_z$ in (\ref{eq:qz}), which is considered by \cite{celis2019classification},
	includes various performance functions used in fairness AI.
	We summarize representative group performance functions having the form of (\ref{eq:qz}) in Table \ref{tab:qz}.
	
	For given group performance functions $q_z(\cdot), z\in \{0,1\},$  we say that $f$ satisfies the BGF constraint with respect to $q_z$ if
	$q_0(f) = q_1(f).$ A relaxed version of  the BGF constraint so called the $\epsilon$-BGF constraint, is frequently considered,
	which requires $|q_0(f)-q_1(f)| < \epsilon$ for a given $\epsilon>0.$
	Typically, AI algorithms search an optimal function $f$ among those satisfying the $\epsilon$-BGF constraint with respect to given group performance functions $q_z(\cdot), z\in \{0,1\}.$
	
	\begin{table*}[t]
		\caption{Some group performance functions}\label{tab:qz}
		\begin{center}
			\begin{footnotesize}
				\begin{sc}
					\begin{tabular}{@{}ccc@{}}
						\toprule
						Fairness criteria & $\mathcal{E}$ & $\mathcal{E}'$  \\
						\midrule
						\begin{tabular}[c]{@{}c@{}} Disparate impact \\ \cite{barocas2016big}\end{tabular}  & $\mathds{1}\{C_f(X)=1\}$ & $\emptyset$  \\
						\begin{tabular}[c]{@{}c@{}}Equal opportunity \\ \cite{hardt2016equality}\end{tabular}  & $\mathds{1}\{C_f(X)=1\}$  & $\{Y=1\}$  \\
						\begin{tabular}[c]{@{}c@{}}Disparate mistreatment w.r.t. Error rate \\ \cite{zafar2019fairness}\end{tabular} & $\mathds{1}\{C_f(X)\neq Y\}$ & $\emptyset$  \\
						\begin{tabular}[c]{@{}c@{}}Mean score parity \\ \cite{coston2019fair}\end{tabular}     & $f(X)$ & $\emptyset$  \\
						\bottomrule
					\end{tabular}
				\end{sc}
			\end{footnotesize}
		\end{center}
	\end{table*}

	\subsection{Related works}
	
	Several learning algorithms have been proposed to find an accurate model $f$ satisfying a given BGF constraint, which are categorized into three groups. 
	In this subsection, we review some methods for each group. 
	\medskip
	
	\textbf{Pre-processing methods:} Pre-processing methods remove bias in training data or find a fair representation with respect to sensitive variables before the training phase and learn AI models based on de-biased data or fair
	representation  \cite{kamiran2012data, zemel2013learning, feldman2015certifying, calmon2017optimized, dixon2018measuring, webster2018mind, xu2018fairgan, creager2019flexibly, quadrianto2019discovering}.
	\cite{kamiran2012data} suggested pre-processing methods to eliminate bias
	in training data by use of label changing, reweighing and sampling. 
	Based on the idea that transformed data should not be able to predict the sensitive variable, \cite{feldman2015certifying} proposed a transformation of 
	input variables for eliminating the disparate impact.
	To find a fair representation, \cite{zemel2013learning, calmon2017optimized}  proposed a data transformation mapping for preserving accuracy and alleviating discrimination simultaneously.
	Pre-processing methods for fair learning on text data were studied by \cite{dixon2018measuring, webster2018mind}.
	\medskip
	
	\textbf{In-processing methods:}
	In-processing methods generally train an AI model by minimizing a given cost function (e.g. the cross-entropy, the sum of squared residuals, the empirical AUC etc.)
	subject to a $\epsilon$-BGF constraint. 
	Most group performance functions $q_z(\cdot)$ are not differentiable, and thus
	various surrogated group performance functions and corresponding $\epsilon$-BGF constraints have been proposed \cite{kamishima2012fairness, goh2016satisfying, bechavod2017learning, zafar2017fairness, zafar2019fairness, donini2018empirical, menon2018cost, narasimhan2018learning, celis2019classification, vogel2020learning, cho2020fair}. 
	\cite{kamishima2012fairness} used a fairness regularizer which is an approximation of the mutual information between the sensitive variable and the target variable. 
	\cite{zafar2017fairness, zafar2019fairness} proposed covariance-type fairness constraints as  tractable proxies targeting the disparate impact and the equality of the false positive or negative rate, and
	\cite{donini2018empirical} used a linear surrogated group performance function
	for the equalized odds.
	On the other hand, \cite{menon2018cost, celis2019classification} derived an optimal classifier for a constrained fair classification as a form of an instance-dependent threshold. 
	Also, for fair score functions, \cite{vogel2020learning}  proposed fairness constraints based on ROC curves of each sensitive group.  
	\medskip
	
	\textbf{Post-processing methods:}
	Post-processing methods first learn an AI model without any BGF constraint and then
	transform the decision boundary or score function of the trained AI model for each sensitive group to satisfy given BGF criteria \cite{kamiran2012decision, fish2016confidence, hardt2016equality, corbett2017algorithmic, pleiss2017fairness, chzhen2019leveraging, wei20a, jiang2020wasserstein}.
	\cite{hardt2016equality, chzhen2019leveraging} suggested finding sensitive group dependent thresholds to get a fair classifier with respect to equal opportunity.
	\cite{wei20a, jiang2020wasserstein} developed an algorithm to transform the original score function to achieve a BGF constraint.

	\section{Within-group fairness for classifiers}
	\label{wgf-classifier}
	
	We assume that there exists a known optimal classifier $C^\star$ which could be the Bayes classifier or its estimate. For example, we
	can use $C_{f^\star}$ for $C^\star,$ where $f^\star$ is the unconstrained minimizer of the cross-entropy on $\cF.$
	We mostly focus on in-processing methods for the BGF and explain how to reflect
	WGF into a learning procedure.  Remarks about how to reflect WGF to
	pre- and post-processing methods are given in Section \ref{remarks}.

	\subsection{Definition of within-group fairness}
	\label{sub-definition}

	Conceptually, WGF means that the classifier $C_f$ and $C^{\star}$ have the same ranks in each sensitive group.
	That is, for two individuals $\bx_A$ and $\bx_B$ in a same sensitive group with $C^\star(\bx_A) > C^\star(\bx_B),$
	WGF requires that $C_f(\bx_A)\ge C_f(\bx_B).$ 
	To materialize this concept of WGF, we define the WGF constraint as
	\be
	\label{eq:wgf}
	\begin{split}
		&\Pr\left\{C^\star(\bX)=0, C_f(\bX)=1 \vert Z=z\right\}=0 \\
		&\mbox{or } \Pr\left\{C^\star(\bX)=1, C_f(\bX)=0 \vert Z=z\right\}=0
	\end{split}
	\ee
	for each $z\in \{0,1\}.$ 
	Similar to the BGF, we relax the constraint (\ref{eq:wgf}) by requiring that
	either of the two probabilities is small. That is, we say that $f$ satisfies the $\delta$-WGF constraint for a given $\delta>0$ if
	\be
	\label{eq:wgf-d}
	\max_{z\in \{0,1\}} \min \{ a_{01 \vert z}(f), a_{10 \vert z}(f) \} <\delta,
	\ee
	where
	$a_{ij \vert z}(f) = \Pr\{C^\star(X)=i, C_f(X)=j \vert Z=z\}.$

	\subsection{Directional within-group fairness}
	\label{sub-directional}
	Many BGF constraints have their own implicit directions toward which the classifier is expected to be guided in the training phase.
	We can design a special WGF constraint reflecting the implicit direction of a given BGF constraint which results
	in more desirable classifiers (better guided, more fair and frequently more accurate). Below, we present two such WGF constraints.
	\medskip
	
	\noindent{\bf Disparate impact:}
	Note that the disparate impact  requires that
	$$\Pr\{C_f(\bX)=1 \vert Z=0\}=\Pr\{C_f(\bX)=1 \vert Z=1\}.$$
	Suppose that $\Pr\{C^\star(\bX)=1 \vert Z=0\}< \Pr\{C^\star(\bX)=1 \vert Z=1\}.$
	Then, we expect that a desirable classifier $C_f$  achieves this BGF constraint by increasing $\Pr\{C_f(\bX)=1 \vert Z=0\}$ from $\Pr\{C^\star(\bX)=1 \vert Z=0\}$
	and decreasing $\Pr\{C_f(\bX)=1 \vert Z=1\}$ from $\Pr\{C^\star(\bX)=1 \vert Z=1\}.$ 
	To reflect this direction, we can enforce a learning algorithm to search a classifier $C_f$ satisfying
	$\Pr\{C^\star(\bX)=1 \vert Z=0\} < \Pr\{C_f(\bX)=1 \vert Z=0\}$ and $\Pr\{C^\star(\bX)=1 \vert Z=1\} > \Pr\{C_f(\bX)=1 \vert Z=1\}.$
	Based on this argument, 
	we define the directional $\delta$-WGF constraint for the disparate impact as
	\be
	\label{eq:dwgf-d}
	\max \{ a_{10 \vert 0}(f), a_{01 \vert 1}(f) \} < \delta.
	\ee
	\medskip
	
	\noindent{\bf Equal opportunity:} The equal opportunity constraint is given as
	\begin{equation*}
		\Pr\{C_f(\bX)=1 \vert Z=0, Y=1\} =\Pr\{C_f(\bX)=1 \vert Z=1, Y=1\}.
	\end{equation*}
	Suppose that $\Pr\{C^\star(\bX)=1 \vert Z=0, Y=1\}< \Pr\{C^\star(\bX)=1 \vert Z=1, Y=1\}.$
	A similar argument for the disparate impact leads us to define the directional $\delta$-WGF constraint for the equal opportunity as
	\be
	\label{eq:dwgf-d-eo-1}
	\max\{ a_{10 \vert 01}(f), a_{01 \vert 11}(f) \} < \delta
	\ee
	and
	\be
	\label{eq:dwgf-d-eo-0}
	\max_{z\in \{0,1\}} \min\left\{ a_{10 \vert z0}(f), a_{01 \vert z0}(f) \right\} < \delta,
	\ee
	where
	$$a_{ij \vert zy}(f) = \Pr\{C^\star(X)=i, C_f(X)=j  \vert  Z=z, Y=y\}.$$
	\medskip
	
	\subsection{Learning with doubly-group fairness constraints}
	
	We say that $f$ satisfies the $(\epsilon,\delta)$-doubly-group fairness constraint if $B(f) < \epsilon$ and $W(f) < \delta,$
	where $B$ is a given BGF constraint and $W$ is the corresponding WGF constraint proposed in the previous two subsections.
	In this section, we propose a relaxed version of $W(\cdot)$ for easy computation. As we review in Section \ref{review}, 
	many relaxed versions of $B(\cdot)$ have been proposed already. 
	
	The WGF constraints considered in Sections \ref{sub-definition} and \ref{sub-directional}
	are hard to be used as themselves in the training phase since
	they are neither convex nor continuous. A standard approach to resolve this problem is to use a convex surrogated function.
	For example, a surrogated version of
	the WGF constraint \eqref{eq:wgf-d} is $W_\text{surr}(f)<\delta,$ where
	\be
	\label{eq:wgf-d-c}
	\begin{split}
		W_\text{surr}(f):= 
		\max_{z\in \{0,1\}} \min \Big\{ &\E\left\{\phi(-f(\bX)) \vert Z=z, Y^\star=1\right\} p_{1 \vert z}, \\ 
		& \E\left\{\phi(f(\bX)) \vert Z=z,Y^\star=0\right\}p_{0 \vert z} \Big\},
	\end{split}
	\ee
	where $Y^\star=C^\star(\bX), p_{y \vert z}=\Pr(C^\star(\bX)=y \vert Z=z)$ and $\phi$ is a convex surrogated function of the indicator function $\mathds{1}(z\ge 0).$ In this paper, we use
	the hinge function given as $\phi_\text{hinge}(z)= (1+z)_{+}$ as a convex surrogated function
	which is popularly used for fair AI \cite{goh2016satisfying, donini2018empirical, wu2018fairness}. 
	The surrogated versions for the other WGF constraints are derived similarly. 
	Finally, we estimate $f$ by $\hat{f}$ that minimizes
	the regularized cost function 
	\be
	\label{eq:cost}
	\cL(f)+\lambda B_\text{surr}(f) + \eta W_\text{surr}(f),
	\ee
	where $\cL$ is a given cost function (e.g. the cross-entropy) and
	$B_\text{surr}$ and $W_\text{surr}$ are the surrogated constraints of $B$ and $W,$ respectively.
	The nonnegative constants $\lambda$ and $\eta$ are regularization parameters which
	are selected so that $\hat{f}$ satisfies 
	$B(\hat{f})<\epsilon$ and $W(\hat{f})<\delta.$

	\subsection{Related notions with within-group fairness}
	
	There are several fairness concepts which are somehow related to WGF. 
	However, the existing concepts are quite different from our WGF.
	\begin{enumerate}
		\item Unified fairness:  \cite{speicher2018unified} used the term `within-group fairness'. 
		However, WGF of \cite{speicher2018unified}
		is different from our WGF. 
		\cite{speicher2018unified} measured individual-level benefits of a given prediction model and they defined the model to be WGF if the individual benefits in each group are similar. 
		They also illustrated that WGF keeps decreasing as
		BGF increases. 
		Our WGF is nothing to do with individual-level benefits. Our WGF can be high even
		when individual-level benefits are not similar. 
		Also, our WGF can increase even when BGF increases. 
		
		\item Slack consistency: \cite{nachum2019group} proposed the `slack consistency' which requires that
		the estimated scores of each individual should be monotonic with respect to slack variables used in fairness constraints.
		Slack consistency does not guarantee within-group fairness because the ranks of the estimated scores can  change even when they move monotonically. 
		
	\end{enumerate}
	
	\section{Within-group fairness for score functions}
	\label{wgf-score}
	
	Similarly to classifiers, the WGF for score functions
	requires that $f(\bx_A) > f(\bx_B)$ when $f^\star(\bx_A)> f^{\star}(\bx_B)$ and vice versa 
	for two individuals $\bx_A$ and $\bx_B$ in a same sensitive group, where
	$f^\star$ is a known optimal score function such as the conditional class probability $\Pr(Y=1 \vert \bX)$
	or its estimate.
	To realize this concept, we define the WGF constraint for a score function $f$ as
	$\tau_z(f)=1$ for $z\in \{0,1\},$ where $\tau_z(\cdot)$ is the Kendall's $\tau$ between $f$ and $f^\star$ conditional on $Z=z,$ that is
	\begin{equation*}
		\tau_z(f)=\E_{(\bX_1,\bX_2)} \Big[ \mathds{1} \{ (f(\bX_1)-f(\bX_2)) (f^\star(\bX_1)-f^\star(\bX_2)) >0\}\Big  \vert  Z=z \Big],
	\end{equation*}
	where $\bX_1$ and $\bX_2$ are independent copies of $\bX.$ In turn, the $\delta$-WGF constraint for a score function $f$ is $1-\tau_z(f) < \delta, z\in \{0,1\}.$
	
	Similarly for classifiers, we need a convex surrogated version of the $\delta$-WGF constraint and a candidate 
	would be $1-\tau_{\phi,z}(f) < \delta, z\in \{0,1\},$ where 
	\begin{equation*}
		\tau_{\phi,z}(f)= 1- \E_{(\bX_1,\bX_2)} \Big[ \phi\{ (f(\bX_1) - f(\bX_2))
		(f^\star(\bX_1) - f^\star(\bX_2)) \} \Big  \vert  Z = z \Big]
	\end{equation*}
	and $\phi$ is a convex surrogated function of $\mathds{1}(z>0)$ such as the $\phi_\text{hinge}.$

	\section{Numerical studies}
	\label{simulation}
	
	We investigate the impacts of the WGF constraints on the prediction accuracy as well as the 
	BGF by analyzing real-world datasets.  
	We consider linear logistic and  deep neural network (DNN) models for $\cF$ and
	use the cross-entropy for $\cL.$ 
	For DNN, fully connected neural networks with one hidden layer and $p$ many hidden nodes are used.
	We train the models by the gradient descent algorithm \cite{bottou2010large} implemented by Python with related libraries \texttt{pytorch, scikit-learn, numpy}.
	The SGD optimizer is used with momentum 0.9 and a learning rate of either  0.1 or 0.01 depending on the dataset. 
	We use the unconstrained minimizer of $\cL$ for $f^\star.$
	\medskip
	
	\textbf{Datasets.} 
	We analyze four real world datasets, which are popularly used in fairness AI research and publicly available: (i) The Adult Income dataset (\textit{Adult}, \cite{Dua:2019}); (ii) The Bank Marketing dataset(\textit{Bank}, \cite{Dua:2019}); (iii) The Law School dataset (\textit{LSAC},  \cite{wightman1998lsac}); (iv) The Compas Propublica Risk Assessment dataset (\textit{COMPAS},  \cite{larson2016we}).
	Except for the dataset \textit{Adult}, we split the training and test datasets randomly by  8:2 ratio and repeat 5 times training/test splits for performance evaluation.

	\subsection{Within-group fair classifiers}
	We consider following group performance functions for the BGF:
	the disparate impact (DI) \cite{barocas2016big} and the disparate mistreatment w.r.t. error rate \cite{zafar2019fairness}, which are defined as
	
	\begin{equation*}
		\begin{split}
			\text{DI}(f) &=  \left\vert \Pr(C_f(\bX)=1  \vert Z=1)
			- \Pr(C_f(\bX)=1 \vert Z=0) \right\vert \\
			\text{ME}(f) &=  \left\vert \Pr(C_f(\bX) \neq Y \vert Z=0)
			- \Pr(C_f(\bX) \neq Y \vert Z=1) \right\vert .
		\end{split}
	\end{equation*}
	Note that the DI is directional while the ME is not.
	For the surrogated BGF constraints, we replace the indicator function with the hinge function in calculating the BGF constraints as is done by \cite{goh2016satisfying, wu2018fairness}. 
	We name the corresponding BGF constraints 
	by Hinge-DI and Hinge-ME respectively.
	The results for other surrogated constraints such as the covariance type constraints proposed by \cite{zafar2017fairness, zafar2019fairness} and the linear surrogated functions considered in \cite{Padala2020FNNC} are presented in the Supplementary material.
	In addition, the results for the equal opportunity constraint are summarized in the Supplementary material.
	
	For investigating the impacts of WGF on trained classifiers, we 
	first fix the $\epsilon$ for each BGF constraint, and we 
	choose the regularization parameters $\lambda$ and $\eta$ to make
	the classifier $\hat{f}$ minimizing the regularized cost function (\ref{eq:cost}) satisfy the $\epsilon$-BGF constraint. 
	Then, we assess the prediction accuracy and the degree of WGF of $\hat{f}.$

	\subsubsection{Targeting for disparate impact}
	
	Table \ref{tab:dnn-table} presents the three $2\times 2$ tables
	comparing the results of the unconstrained DNN classifier ($\hat{Y}^\star$)  and  three DNN classifiers ($\hat{Y}$)  trained on the dataset \textit{Adult}:
	(i) only with the DI constraint, (ii) with the DI and WGF constraints and (iii) with the DI and directional WGF (dWGF) constraints. We let $\epsilon$ be around 0.03.
	The numbers marked in red are subjects treated unfairly with respect to the dWGF.
	Note that the numbers of unfairly treated subjects are reduced much with the WGF and dWGF constraints and the dWGF constraint is more effective. We report that
	the  accuracies of the three classifiers on the test data are
	0.837, 0.840 and 0.839, respectively, which indicates that the WGF and dWGF constraints improve the WGF without hampering the accuracy.
	Compared to the dWGF, the WGF constraint is less effective, which is observed consistently 
	for different datasets when a BGF constraint is directional. See Table
	2 in the Supplementary material for the corresponding numerical results. Thus, hereafter we consider the dWGF only for the DI which has an implicit direction.

	\begin{table}[t]
		\caption{Comparison of the results of the three DNN classifiers trained
			(i) only with the BGF constraint, (ii) with the BGF and WGF constraints and (iii)
			with the BGF and dWGF constraints on the dataset \textit{Adult}.
			Marked in red represent the numbers of subjects treated unfairly
			in a same sensitive group.}
		\label{tab:dnn-table}
		\begin{center}
			\begin{footnotesize}
				\begin{sc}
					\begin{tabular}{cccccc}
						\toprule
						\multicolumn{6}{c}{Only with the DI constraint } \\
						\multicolumn{3}{c}{$Z=0$} & \multicolumn{3}{c}{$Z=1$} \\ \cmidrule(lr){1-3} \cmidrule(lr){4-6}
						& $\hat{Y}=0$ & $\hat{Y}=1$ & & $\hat{Y}=0$ & $\hat{Y}=1$ \\
						$\hat{Y}^\star=0$ & 4,592 & 350 & $\hat{Y}^\star=0$ & 7,966 & \textcolor{red}{86} \\
						$\hat{Y}^\star=1$ & \textcolor{red}{13} & 466 & $\hat{Y}^\star=1$ & 945 & 1,863 \\
						\midrule
						
						\multicolumn{6}{c}{With the DI and WGF constraints}
						\\
						\multicolumn{3}{c}{$Z=0$} & \multicolumn{3}{c}{$Z=1$} \\ \cmidrule(lr){1-3} \cmidrule(lr){4-6}
						& $\hat{Y}=0$ & $\hat{Y}=1$ & & $\hat{Y}=0$ & $\hat{Y}=1$ \\
						$\hat{Y}^\star=0$ & 4,703 & 239 & $\hat{Y}^\star=0$ & 8,021 & \textcolor{red}{31} \\
						$\hat{Y}^\star=1$ & \textcolor{red}{27} & 452 & $\hat{Y}^\star=1$ & 1,156 & 1,652 \\
						\midrule
						
						\multicolumn{6}{c}{With the DI and dWGF constraints}
						\\
						\multicolumn{3}{c}{$Z=0$} & \multicolumn{3}{c}{$Z=1$} \\ \cmidrule(lr){1-3} \cmidrule(lr){4-6}
						& $\hat{Y}=0$ & $\hat{Y}=1$ & & $\hat{Y}=0$ & $\hat{Y}=1$ \\
						$\hat{Y}^\star=0$ & 4,718 & 224 & $\hat{Y}^\star=0$ & 8,024 & \textcolor{red}{28} \\
						$\hat{Y}^\star=1$ & \textcolor{red}{18} & 461 & $\hat{Y}^\star=1$ & 1,178 & 1,630 \\
						\bottomrule
					\end{tabular}
				\end{sc}
			\end{footnotesize}
		\end{center}
	\end{table}

	Table \ref{tab:di-dwgf} summarizes the performances 
	of the three classifiers - $C^\star$ and the two classifiers trained  with the DI constraint and  the DI and dWGF constraints (doubly-fair, DF), respectively. In Table \ref{tab:di-dwgf}, we report the accuracies as well as the values
	of DI and dWGF terms (i.e., $\text{DI}(\hat{f})$ and
	$\max \{ a_{10 \vert 0}(\hat{f}), a_{01 \vert 1}(\hat{f}) \},$ respectively). 
	We observe that the DF classifier improves the dWGF while keeping that the DI values and accuracies are favorably comparable to those of the BGF classifier.
	For reference,  the performances with the WGF constraint are summarized in the Supplementary material.

	\begin{table*}[t]
		\caption{Results for the DF classifier with the Hinge-DI constraint. Except for the dataset \textit{Adult}, the average performances are given.}
		\label{tab:di-dwgf}
		\begin{center}
			\begin{footnotesize}
				\begin{sc}
					\begin{tabular}{llccclccc}
						\toprule
						&          & \multicolumn{3}{c}{Linear model} &  & \multicolumn{3}{c}{DNN model} \\ \cmidrule(lr){3-5} \cmidrule(lr){7-9} 
						Dataset & Method      & ACC   & DI    & dWGF  &  & ACC   & DI    & dWGF  \\ \midrule
						\textit{Adult}  & Uncons.     & 0.852 & 0.172 & 0.000 &  & 0.853 & 0.170 & 0.000 \\
						& Hinge-DI    & 0.833 & 0.028 & 0.005 &  & 0.837 & 0.029 & 0.008 \\
						& Hinge-DI-DF & 0.836 & 0.028 & 0.003 &  & 0.839 & 0.026 & 0.003 \\ \midrule
						\textit{Bank}    & Uncons.     & 0.908 & 0.195 & 0.000 &  & 0.904 & 0.236 & 0.000 \\
						& Hinge-DI    & 0.901 & 0.024 & 0.018 &  & 0.899 & 0.029 & 0.033 \\
						& Hinge-DI-DF & 0.904 & 0.021 & 0.007 &  & 0.905 & 0.029 & 0.032 \\ \midrule
						\textit{LSAC}    & Uncons.     & 0.823 & 0.120 & 0.000 &  & 0.856 & 0.131 & 0.000 \\
						& Hinge-DI    & 0.809 & 0.016 & 0.014 &  & 0.816 & 0.032 & 0.064 \\
						& Hinge-DI-DF & 0.813 & 0.018 & 0.009 &  & 0.809 & 0.029 & 0.047 \\ \midrule
						\textit{COMPAS}  & Uncons.     & 0.757 & 0.164 & 0.000 &  & 0.757 & 0.162 & 0.000 \\
						& Hinge-DI    & 0.641 & 0.024 & 0.153 &  & 0.639 & 0.030 & 0.142 \\
						& Hinge-DI-DF & 0.618 & 0.025 & 0.145 &  & 0.654 & 0.033 & 0.120 \\ 
						
						\bottomrule
					\end{tabular}
				\end{sc}
			\end{footnotesize}
		\end{center}
	\end{table*}

\begin{figure}[ht]
		\begin{center}
			\centerline{\includegraphics[width=0.7\columnwidth]{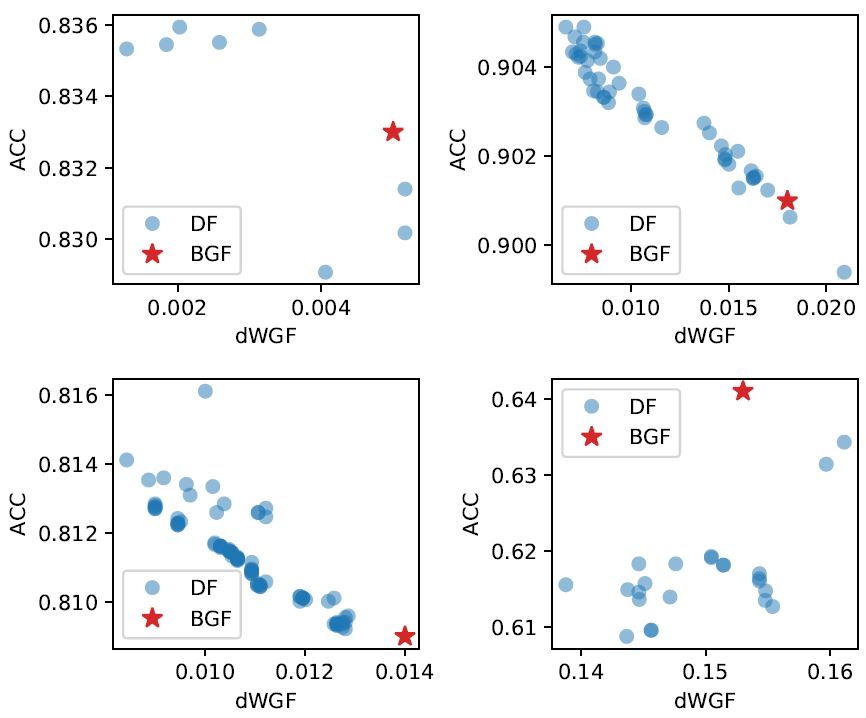}}
			\caption{Scatter plots of the accuracies and dWGF values for the DF linear regression model with the DI values around 0.03. (Topleft) \textit{Adult}; (Topright) \textit{Bank}; (Bottomleft) \textit{LSAC}; (Bottomright) \textit{COMPAS}. Red star points in each figure represent the results of the BGF classifier. }
			\label{fig:lin-dwgf}
		\end{center}
		\vskip -0.2in
	\end{figure}

	To investigate the sensitivity of the accuracy to the degree of WGF, 
	the scatter plots between various dWGF values and the corresponding accuracies 
	for the DF linear logistic model are given in  Figure \ref{fig:lin-dwgf}, where
	the DI value is fixed around 0.03. The accuracies are not sensitive to the dWGF values.
	Moreover, for the datasets \textit{Adult}, \textit{Bank} and \textit{LSAC},
	the accuracies keep increasing as the dWGF value decreases.
	
\begin{figure}[ht]
		\begin{center}
			\centerline{\includegraphics[width=0.7\columnwidth]{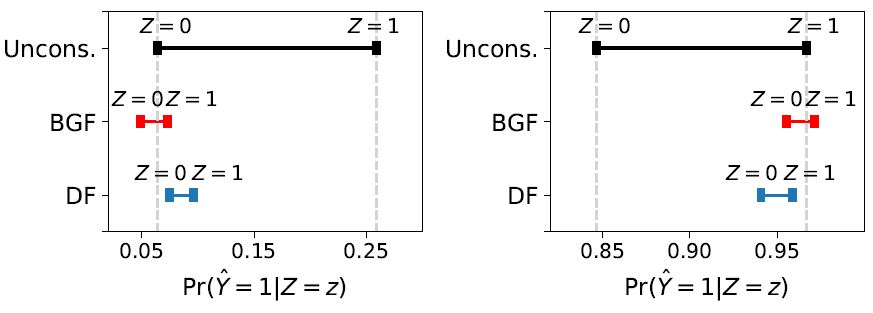}}
			\caption{Comparison of the  conditional probabilities
				of each group for the datasets \textit{Bank} (Left) and \textit{LSAC}(Right).}
			\label{fig:lin-res-phat}
		\end{center}
		\vskip -0.2in
	\end{figure}

	While we analyzed the datasets \textit{Bank} and \textit{LSAC}, we found an undesirable aspect of the learning algorithm only with the DI constraint.
	The corresponding classifiers improve the DI by decreasing (or increasing) the
	probabilities $P(\hat{Y}=1 \vert Z=0)$ and $P(\hat{Y}=1 \vert Z=1)$ simultaneously
	compared to $P(Y^\star=1 \vert Z=0)$ and $P(Y^\star=1 \vert Z=1).$
	A better way to improve the DI would be to increase
	$P(\hat{Y}=1 \vert Z=0)$ and  decrease $P(\hat{Y}=1 \vert Z=1)$ when
	$P(Y^\star=1 \vert Z=0)< P(Y^\star=1 \vert Z=1).$ Figures \ref{fig:lin-res-phat} show that this undesirable aspect disappears when the dWGF constraint is considered.

	\subsubsection{Targeting for disparate mistreatment} 
	
	The results of the performances of the DF classifier with the ME as a BGF constraint are presented in Table \ref{tab:me-wgf}. Since the ME has no implicit direction, we use the undirectional WGF constraint. The overall conclusions are similar to those for the DI and dWGF constraints. That is, the undirectional WGF constraint also works well.
	
	\begin{table*}
		\caption{Results for the DF classifier with the Hinge-ME constraint. Except for the dataset \textit{Adult}, average performances are given.}
		\label{tab:me-wgf}
		\begin{center}
			\begin{footnotesize}
				\begin{sc}
					\begin{tabular}{llccccccc}
						\toprule
						&          & \multicolumn{3}{c}{Linear model} &  & \multicolumn{3}{c}{DNN model} \\ \cmidrule(lr){3-5} \cmidrule(lr){7-9} 
						Dataset         & Method   & ACC      & ME  		 & WGF     &  & ACC     & ME	     & WGF    \\ \midrule
						\textit{Adult} & Uncons.  & 0.852    & 0.117      & 0.000    &  & 0.853   & 0.105     & 0.000   \\
						& Hinge-ME    & 0.834    & 0.060      & 0.005    &  & 0.822   & 0.025     & 0.059   \\
						& Hinge-ME-DF & 0.834    & 0.060      & 0.005    &  & 0.825   & 0.031   & 0.026 \\
						\midrule
						\textit{Bank}   & Uncons.  & 0.908    & 0.177      & 0.000    &  & 0.904   & 0.174     & 0.000   \\
						& Hinge-ME    & 0.740    & 0.044      & 0.068    &  & 0.902   & 0.164     & 0.076   \\
						& Hinge-ME-DF & 0.749    & 0.045      & 0.020    &  & 0.897   & 0.165     & 0.047   \\
						\midrule
						\textit{LSAC}   & Uncons.  & 0.823    & 0.090      & 0.000    &  & 0.856   & 0.071     & 0.000   \\
						& Hinge-ME    & 0.759    & 0.028      & 0.038    &  & 0.815   & 0.044     & 0.040   \\
						& Hinge-ME-DF & 0.742    & 0.020      & 0.017    &  & 0.803   & 0.038     & 0.001   \\
						\midrule
						\textit{COMPAS} & Uncons.  & 0.757    & 0.022      & 0.000    &  & 0.757   &   0.024   &  0.000 \\
						& Hinge-ME    & 0.740    & 0.020      & 0.018    &  & 0.738   &   0.016   &  0.018 \\
						& Hinge-ME-DF & 0.743    & 0.018      & $<$0.001 &  & 0.757   &   0.017   &  0.001 \\
						\bottomrule
					\end{tabular}
				\end{sc}
			\end{footnotesize}
		\end{center}
	\end{table*}

	\subsection{Within-group fair for score function}
	
	In this section, we examine the WGF constraint for score functions. 
	We choose the logistic loss (binary cross-entropy, BCE) and AUC (area under the ROC) as evaluation metrics for prediction accuracy.
	For the BGF, we consider the mean score parity (MSP, \cite{coston2019fair}):
	$$\text{MSP} (f)= \left\vert \E(\sigma(f(X)) \vert Z=1) - \E(\sigma(f(X)) \vert Z=0)\right\vert,$$
	where $\sigma: x \mapsto 1/(1+e^{-x})$ is the sigmoid function.
	To check how much the estimated score function $\hat{f}$ is within-group fair, we calculate the Kendall's $\tau$ between  $\hat{f}$ and the ground-truth score
	function $f^\star$ on the test data for each sensitive group, and then we average them, which is denoted by $\bar{\tau}$ in Table \ref{tab:msp-res}. We choose the regularization parameters $\lambda$ and $\eta$ such that $\bar{\tau}$ 
	of $\hat{f}$ is as close to 1 as possible while maintaining the MSP value around 0.03.
	
	Table \ref{tab:msp-res} amply shows that the DF score function
	always  improves the degree of WGF (measured by $\bar{\tau}$) and the accuracy in terms of AUC simultaneously while keeping the degree of BGF at a reasonable level.
	With respect to the BCE, the BGF and DF score functions are similar.
	The superiority of the DF score function in terms of AUC compared with the BGF score function is partly because the WGF constraint shrinks the estimated score toward the ground-truth score (Uncons. in Table \ref{tab:msp-res}) which is expected to be most accurate. Based on these results, we conclude that the WGF constraint is a useful guide
	to find a better score function with respect to AUC as well as the WGF.

	\begin{table*}[t]
		\caption{Results of the DF score functions. Except for the dataset \textit{Adult}, averages performances are given.}
		\label{tab:msp-res}
		\begin{center}
			\begin{footnotesize}
				\begin{sc}
					\resizebox{\textwidth}{!}{
						\begin{tabular}{llccccccccc}
							\toprule
							&         & \multicolumn{4}{c}{Linear model}     &  & \multicolumn{4}{c}{DNN model}        \\ \cmidrule(lr){3-6} \cmidrule(lr){8-11} 
							Dataset         & Method  & BCE   & AUC   & MSP   & $\bar{\tau}$ &  & BCE   & AUC   & MSP   & $\bar{\tau}$ \\ \midrule
							\textit{Adult} & Uncons. & 0.319 & 0.905 & 0.173 & 1.000        &  & 0.315 & 0.908 & 0.178 & 1.000        \\
							& BGF     & 0.358 & 0.879 & 0.037 & 0.854        &  & 0.353 & 0.879 & 0.035 & 0.805        \\
							& DF      & 0.368 & 0.882 & 0.033 & 0.908        &  & 0.364 & 0.885 & 0.035 & 0.891        \\ \midrule
							\textit{Bank}   & Uncons. & 0.214 & 0.932 & 0.217 & 1.000        &  & 0.237 & 0.926 & 0.237 & 1.000        \\
							& BGF     & 0.235 & 0.906 & 0.036 & 0.706        &  & 0.270 & 0.908 & 0.033 & 0.671        \\
							& DF      & 0.240 & 0.912 & 0.039 & 0.728        &  & 0.266 & 0.917 & 0.031 & 0.761        \\ \midrule
							\textit{LSAC}   & Uncons. & 0.434 & 0.732 & 0.125 & 1.000        &  & 0.359 & 0.831 & 0.142 & 1.000        \\
							& BGF     & 0.450 & 0.705 & 0.033 & 0.692        &  & 0.381 & 0.803 & 0.025 & 0.640        \\
							& DF      & 0.557 & 0.717 & 0.031 & 0.719        &  & 0.383 & 0.809 & 0.028 & 0.738        \\ \midrule
							\textit{COMPAS} & Uncons. & 0.511 & 0.822 & 0.122 & 1.000        &  & 0.506 & 0.824 & 0.118 & 1.000        \\
							& BGF     & 0.599 & 0.759 & 0.035 & 0.564        &  & 0.588 & 0.753 & 0.030 & 0.561        \\
							& DF      & 0.597 & 0.792 & 0.038 & 0.720        &  & 0.597 & 0.766 & 0.028 & 0.623        \\ 
							\bottomrule
						\end{tabular}
					}
				\end{sc}
			\end{footnotesize}
		\end{center}
	\end{table*}

	\section{Remarks on within-group fairness for pre- and post-processing methods}
	\label{remarks}
	
	Various pre- and post-processing methods for fair AI have been proposed.
	An advantage of these methods compared to constrained methods is that the methods are simple, computationally efficient but yet reasonably accurate.
	In this section, we briefly explain how to reflect the WGF to pre- and post-processing methods for the BGF.
	
	\subsection{Pre-processing methods and within-group fairness}
	
	Basically, pre-processing methods transform the training data in a certain way to be between-group fair and train an AI model on the transformed data.
	To reflect the WGF, it suffices to add a WGF constraint in the training phase. 
	Let $\cD_{\text{trans}}$ be the transformed training data to be between-group fair and let
	$\cL_{\text{trans}}$ be the corresponding cost function. 
	Then, we learn a model by minimizing $L_{\text{trans}}(f)+\eta W_{\text{conv}}(f)$
	for $\eta>0.$ 
	
	Table \ref{tab:pre-dwgf} presents the results of the models trained on the pre-processing training data and a WGF constraint for various values of $\eta,$ where the DI is used as the BGF and thus the corresponding dWGF constraint is used.
	In this experiment, we use the linear logistic model and the Massaging \cite{kamiran2012data} for the pre-processing. 
	Surprisingly we observed that introducing the dWGF constraint to the pre-processing method helps to improve the BGF and WGF simultaneously without 
	sacrificing the accuracies much.

	\begin{table}[t]
		\caption{Comparison of the accuracy and fairnesses of the pre-processing method
			with and without the dWGF constraint. The results are evaluated on the dataset \textit{Adult}.}
		\label{tab:pre-dwgf}
		\begin{center}
			\begin{footnotesize}
				\begin{sc}
					\begin{tabular}{lcccc}
						\toprule
						Method & $\eta$ & Acc & DI & dWGF \\
						\midrule
						Massaging    	   & -    & 0.837 & 0.069 & 0.009 \\
						Massaging + dWGF & 0.5  & 0.837 & 0.048 & 0.004 \\
						& 1.0  & 0.836 & 0.037 & 0.003 \\
						\bottomrule
					\end{tabular}
				\end{sc}
			\end{footnotesize}
		\end{center}
	\end{table}

	\subsection{Post-processing methods and within-group fairness}

	For the BGF score functions, \cite{jiang2020wasserstein} developed an algorithm to obtain two monotonically nondecreasing transformations $m_z, z\in\{0,1\}$ such that
	$m_0\circ f^\star$ and $m_1\circ f^\star$ are BGF in the sense that the distributions of $m_0\circ f^\star(\bX)\vert Z=0$ and
	$m_1\circ f^\star(\bX) \vert Z=1$ are the same. It is easy to check that the transformed score function $m_z\circ f^\star(\bx)$ is a perfectly WGF score function even though it depends on the sensitivity group variable $z.$ 
	Note that the algorithm in Section \ref{wgf-score} yields score functions not depending on $z.$

	\section{Conclusion}
	\label{conclusion}

	In this paper, we introduced a new concept so called {\it within-group fairness}, which should be considered
	along with BGF when fair AI is a concern. Also, we proposed a regularization procedure to 
	control the degree of WGF of the estimated classifiers and score functions.
	By analyzing four real-world datasets, we illustrated that the WGF constraints
	improve the degree of WGF without hampering
	BGF as well as accuracy. Moreover, in many cases, the WGF constraints are
	helpful to find more accurate prediction models. 
	
	A problem in the proposed learning algorithm for WGF is that
	using a surrogated constraint for a given WGF constraint
	is sometimes problematic. The learning algorithm can find a DF model which
	has a lower surrogated WGF value than that of a BGF model, but the original WGF value is much higher. See Section A.2 of Appendix for empirical evidence. 
	A better surrogated WGF constraint to ensure a lower original WGF value 
	would be useful.

	\section*{Acknowledgments}
	This work was supported by Institute for Information \& communications Technology Planning \& Evaluation(IITP) grant funded by the Korea government(MSIT) (No. 2019-0-01396, Development of framework for analyzing, detecting, mitigating of bias in AI model and training data).

	\bibliography{WGF.bib}
	\bibliographystyle{unsrt}  
	 
	\newpage
	\appendix
	
	\section{Supplenmetary Material}
	\subsection{Additional numerical studies for WGF classification}

	\subsubsection{Targeting for disparate impact} 
	
\begin{figure}[h]
		\begin{center}
			\centerline{\includegraphics[width=0.7\columnwidth]{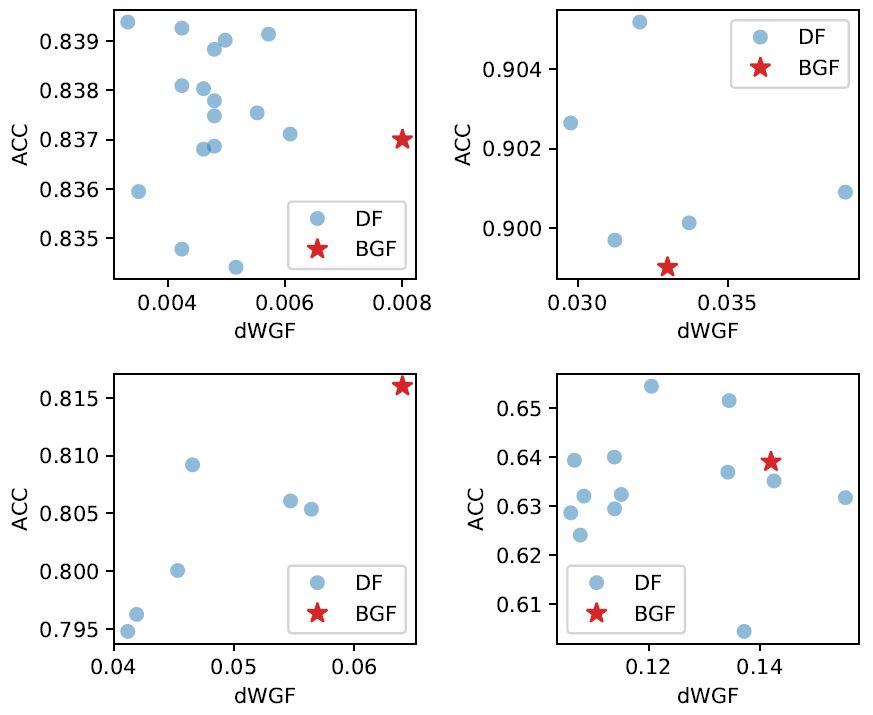}}
			\caption{Scatter plots of the accuracies and dWGF values for the
				DF DNN model with the DI values around 0.03. (Topleft) \textit{Adult}; (Topright) \textit{Bank}; (Bottomleft) \textit{LSAC}; (Bottomright) \textit{COMPAS}. Red star points in each figure represent the results of the BGF classifier. }
			\label{fig:app-dnn-dwgf}
		\end{center}
		\vskip -0.2in
	\end{figure}

	
	First, we investigate the sensitivity of the prediction accuracy to the degree of dWGF in the DNN model.
	Figure \ref{fig:app-dnn-dwgf} shows the scatter plots between various dWGF values and the corresponding accuracies for the DF DNN model, where the DI is fixed around 0.03. 
	The accuracies are not very sensitive to the dWGF values like the DF linear logistic model.
	Furthermore, for the datasets \textit{Adult}, \textit{Bank} and \textit{COMPAS}, the DF classifiers have higher accuracies and lower dWGF values than the BGF classifier.
	
	We also investigate how the dWGF constraint performs with surrogated BGF constraints other than Hinge-DI: (i) the covariance type constraint \cite{zafar2017fairness, zafar2019fairness}, named by COV-DI; 
	and (ii) the linear surrogated function, named by FNNC-DI \cite{Padala2020FNNC}.
	Table \ref{tab:app-di-dwgf} presents the results with various surrogated DI constraints and the dWGF constraint.
	In most cases, COV-DI and FNNC-DI give the results similar to Hinge-DI with or without the dWGF constraint and we consistently observe that considering the dWGF constraint together with the DI constraint helps to alleviate within-group fairness while maintaining similar levels of the accuracy and the DI.
	Note that for the dataset \textit{Adult}, the DNN model with COV-DI constraint does not achieve the pre-specified DI value 0.03 regardless of the choice of tuning parameter.
	In contrast, the DNN model trained with the DI and dWGF constraints 
	achieves the DI value 0.03 with a smaller value of dWGF. 
	This observation is interesting since it implies that the dWGF constraint is helpful to increase even the BGF.
	
	\begin{table}[ht]
		\caption{Results for the DF classifier with various surrogated DI constraints. Except for the dataset \textit{Adult}, average performances are described.}
		\label{tab:app-di-dwgf}
		\begin{center}
			\begin{footnotesize}
				\begin{sc}
					\begin{tabular}{llccclccc}
						\toprule
						&          & \multicolumn{3}{c}{Linear model} &  & \multicolumn{3}{c}{DNN model} \\ \cmidrule(lr){3-5} \cmidrule(lr){7-9} 
						Dataset & Method      & ACC   & DI    & dWGF  &  & ACC   & DI    & dWGF  \\ \midrule
						\textit{Adult}  & Uncons.     & 0.852 & 0.172 & 0.000 &  & 0.853 & 0.170 & 0.000 \\
						& COV-DI      & 0.837 & 0.035 & 0.003 &  & 0.845 & 0.082 & 0.013 \\
						& COV-DI-DF   & 0.837 & 0.030 & 0.001 &  & 0.840 & 0.025 & 0.007 \\
						& FNNC-DI     & 0.834 & 0.023 & 0.003 &  & 0.838 & 0.023 & 0.006 \\
						& FNNC-DI-DF  & 0.836 & 0.025 & 0.001 &  & 0.841 & 0.025 & 0.004 \\ \midrule
						\textit{Bank}    & Uncons.     & 0.908 & 0.195 & 0.000 &  & 0.904 & 0.236 & 0.000 \\
						& COV-DI      & 0.904 & 0.019 & 0.009 &  & 0.906 & 0.019 & 0.036 \\
						& COV-DI-DF   & 0.904 & 0.020 & 0.007 &  & 0.906 & 0.020 & 0.033 \\
						& FNNC-DI     & 0.903 & 0.020 & 0.013 &  & 0.901 & 0.020 & 0.029 \\
						& FNNC-DI-DF  & 0.905 & 0.020 & 0.008 &  & 0.900 & 0.010 & 0.027 \\ \midrule
						\textit{LSAC}    & Uncons.     & 0.823 & 0.120 & 0.000 &  & 0.856 & 0.131 & 0.000 \\
						& COV-DI      & 0.808 & 0.015 & 0.014 &  & 0.859 & 0.052 & 0.020 \\
						& COV-DI-DF   & 0.811 & 0.019 & 0.010 &  & 0.860 & 0.054 & 0.014 \\
						& FNNC-DI     & 0.809 & 0.020 & 0.014 &  & 0.851 & 0.025 & 0.023 \\
						& FNNC-DI-DF  & 0.809 & 0.014 & 0.010 &  & 0.844 & 0.010 & 0.019 \\ \midrule
						\textit{COMPAS}  & Uncons.     & 0.757 & 0.164 & 0.000 &  & 0.757 & 0.162 & 0.000 \\
						& COV-DI      & 0.640 & 0.029 & 0.149 &  & 0.661 & 0.038 & 0.124 \\
						& COV-DI-DF   & 0.620 & 0.024 & 0.135 &  & 0.650 & 0.028 & 0.097 \\
						& FNNC-DI     & 0.646 & 0.037 & 0.146 &  & 0.646 & 0.032 & 0.133 \\
						& FNNC-DI-DF  & 0.624 & 0.034 & 0.143 &  & 0.645 & 0.021 & 0.117 \\ 
						\bottomrule
					\end{tabular}
				\end{sc}
			\end{footnotesize}
		\end{center}
	\end{table}

	Next, we compare the dWGF and WGF constraints when targeting the DI with the hinge surrogated function in Table \ref{tab:app-di-wgf}.
	In most cases, both the dWGF and WGF constraints are helpful to improve the WGF, while maintaining a similar level of accuracy and DI.
	It is noticeable that the DF classifier with the dWGF constraint is more accurate than that with the WGF constraint, which would be mainly because the DI constraint is directional.
	
	\begin{table}[t]
		\caption{Comparison of the dWGF and WGF constraints based on the linear logistic model. Except for the dataset \textit{Adult}, average performances are described.}
		\label{tab:app-di-wgf}
		\begin{center}
			\begin{footnotesize}
				\begin{sc}
					\resizebox{\textwidth}{!}{
						\begin{tabular}{llccclccc}
							\toprule
							
							&             & \multicolumn{3}{c}{with the dWGF constraint} &  & \multicolumn{3}{c}{with the WGF constraint} \\ \cmidrule(lr){3-5} \cmidrule(lr){7-9} 
							Dataset & Method      & ACC           & DI            & dWGF          &  & ACC           & DI            & WGF           \\ \midrule
							\textit{Adult}  & Hinge-DI    & 0.833         & 0.028         & 0.005         &  & 0.833         & 0.028         & 0.005         \\
							& Hinge-DI-DF & 0.836         & 0.028         & 0.003         &  & 0.830         & 0.012         & 0.005         \\ \midrule
							\textit{Bank}    & Hinge-DI    & 0.901         & 0.024         & 0.018         &  & 0.901         & 0.024         & 0.003         \\
							& Hinge-DI-DF & 0.904         & 0.021         & 0.007         &  & 0.898         & 0.017         & 0.000         \\ \midrule
							\textit{LSAC}    & Hinge-DI    & 0.809         & 0.017         & 0.014         &  & 0.809         & 0.017         & 0.014         \\
							& Hinge-DI-DF & 0.813         & 0.018         & 0.009         &  & 0.810         & 0.016         & 0.011         \\ \midrule
							\textit{COMPAS}  & Hinge-DI    & 0.641         & 0.024         & 0.153         &  & 0.641         & 0.024         & 0.136         \\
							& Hinge-DI-DF & 0.618         & 0.025         & 0.145         &  & 0.594         & 0.018         & 0.088         \\
							\bottomrule
						\end{tabular}
					}
				\end{sc}
			\end{footnotesize}
		\end{center}
	\end{table}

	\subsubsection{Targeting for equal opportunity}
	\label{numerical-EOp} 
	We exam how the dWGF constraint works with the equal opportunity constraint given as
	\begin{equation*}
		\begin{split}
			\text{EOp} &= \left\vert \Pr(\hat{Y}=1 \vert Y = 1, Z=1)- \Pr(\hat{Y}=1 \vert Y = 1, Z=0)\right\vert,
		\end{split}
	\end{equation*}
	and the results are summarized in Table \ref{tab:app-eop-dwgf}.
	For some cases, the dWGF constraint does not work at all (i.e., the dWGF values of the BGF and DF classifiers are the sames). 
	This is partly because the surrogated dWGF constraint does not represent the original dWGF well, which is discussed in the following section.
	
	\begin{table*}
		\caption{Results for targeting EOp-dWGF. Except for the dataset \textit{Adult}, average performances are described.}
		\label{tab:app-eop-dwgf}
		\begin{center}
			\begin{footnotesize}
				\begin{sc}
					\begin{tabular}{llccclccc}
						\toprule
						&          & \multicolumn{3}{c}{Linear model} &  & \multicolumn{3}{c}{DNN model} \\ \cmidrule(lr){3-5} \cmidrule(lr){7-9} 
						Dataset & Method       & ACC   & EOp   & dWGF  &  & ACC   & EOp   & dWGF  \\ \midrule
						\textit{Adult}  & Uncons.      & 0.852 & 0.070 & 0.000 &  & 0.853 & 0.076 & 0.000 \\
						& Hinge-EOp    & 0.851 & 0.011 & 0.002 &  & 0.854 & 0.012 & 0.030 \\
						& Hinge-EOp-DF & 0.853 & 0.016 & 0.001 &  & 0.854 & 0.015 & 0.012 \\
						& FNNC-EOp     & 0.851 & 0.013 & 0.012 &  & 0.852 & 0.004 & 0.021 \\
						& FNNC-EOp-DF  & 0.852 & 0.007 & 0.007 &  & 0.852 & 0.006 & 0.019 \\ \midrule
						\textit{Bank}    & Uncons.      & 0.908 & 0.099 & 0.000 &  & 0.904 & 0.082 & 0.000 \\
						& Hinge-EOp    & 0.908 & 0.027 & 0.007 &  & 0.909 & 0.031 & 0.122 \\
						& Hinge-EOp-DF & 0.908 & 0.027 & 0.007 &  & 0.909 & 0.031 & 0.122 \\
						& FNNC-EOp     & 0.908 & 0.027 & 0.010 &  & 0.903 & 0.037 & 0.111 \\
						& FNNC-EOp-DF  & 0.908 & 0.030 & 0.010 &  & 0.900 & 0.028 & 0.107 \\ \midrule
						\textit{LSAC}    & Uncons.      & 0.823 & 0.041 & 0.000 &  & 0.856 & 0.038 & 0.000 \\
						& Hinge-EOp    & 0.820 & 0.003 & 0.004 &  & 0.852 & 0.010 & 0.015 \\
						& Hinge-EOp-DF & 0.820 & 0.003 & 0.004 &  & 0.851 & 0.008 & 0.012 \\
						& FNNC-EOp     & 0.822 & 0.011 & 0.003 &  & 0.859 & 0.010 & 0.011 \\
						& FNNC-EOp-DF & 0.822 & 0.011 & 0.003 &  & 0.858 & 0.010 & 0.010 \\ \midrule
						\textit{COMPAS}  & Uncons.      & 0.757 & 0.074 & 0.000 &  & 0.757 & 0.075 & 0.000 \\
						& Hinge-EOp    & 0.713 & 0.042 & 0.073 &  & 0.719 & 0.029 & 0.046 \\
						& Hinge-EOp-DF & 0.713 & 0.042 & 0.073 &  & 0.719 & 0.029 & 0.046 \\
						& FNNC-EOp     & 0.666 & 0.039 & 0.197 &  & 0.722 & 0.031 & 0.056 \\
						& FNNC-EOp-DF  & 0.706 & 0.031 & 0.092 &  & 0.725 & 0.035 & 0.042 \\ 
						\bottomrule
					\end{tabular}
				\end{sc}
			\end{footnotesize}
		\end{center}
	\end{table*}

	\subsection{Limitations of surrogated WGF constraint}
	\label{limitations}
	
\begin{figure}[h]
		\begin{center}
			\centerline{\includegraphics[width=0.95\columnwidth]{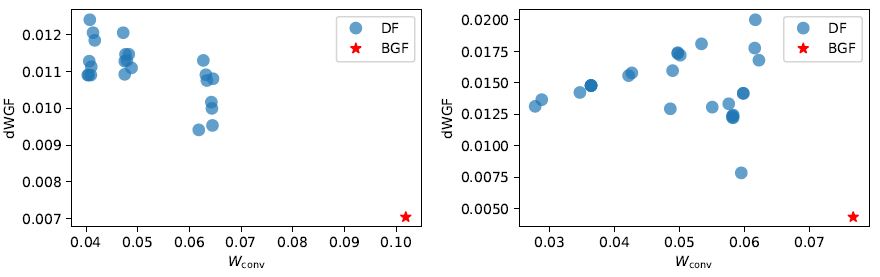}}
			\caption{Scatter plots of the dWGF and the within-group fairness penalty ($W_\text{conv}$) values for the DF linear logistic model with the EOp values around 0.03 evaluated on the training datasets. (Left) \textit{Bank}; (Right) \textit{LSAC}. Red star points in each figure represent the results of the BGF classifier. }
			\label{fig:limit-surr}
		\end{center}
		\vskip -0.2in
	\end{figure}

	We have seen that the DF classifier does not improve the dWGF value at all compared to the BGF classifier with respect to the equal opportunity constraint for some datasets. 
	We found that these undesirable results would be because the surrogated dWGF constraint using the hinge function does not represent the original dWGF constraint.
	To take a closer look at this problem, we investigate relations between the dWGF and $W_\text{conv}$ evaluated on the training datasets \textit{Bank} and \textit{LSAC} in Figure \ref{fig:limit-surr}.
	We observe that the DF classifier has lower $W_\text{conv}$ values but higher dWGF values than the BGF classifier.
	That is, reducing the $W_\text{conv}$ value does not always result in a small value of the original dWGF. 
	Alternative surrogated constraints, which resemble the original dWGF closely but are yet computationally easy, are needed and we leave this issue for future work.
	
	\subsection{Datasets and Preprocessing}
	\textbf{Dataset.} 
	We conduct our experiments with four real-world datasets, which are popularly used in fairness AI research and publicly available: 
	\begin{itemize}
		
		\item \textit{Adult} \cite{Dua:2019}: The Adult Income dataset consists of 32,561 training subjects and 16,281 test subjects with 14 features and a binary target, which indicates whether income exceeds \$50k per a year. The sensitive variable is the sex of the subject,  $Z=0$ for female and $Z=1$ for male.
		
		\item \textit{Bank} \cite{Dua:2019}: The Bank Marketing dataset contains 41,188 subjects with 20 features (e.g. age, occupation, marital status) and a binary target indicating whether or not subjects have subscribed to the product (bank term deposit). A discrete age is set as a binary sensitive variable by assigning 0 to subjects aged 25 to 60 years old and 1 to else.
		
		\item \textit{LSAC} \cite{wightman1998lsac}: The Law School dataset pre-processed by  \cite{lahoti2020fairness} contains 26,551 subjects with 10 input variables and a binary target which indicates whether subject passed the bar exam or not. 
		The sensitive variable is set by 0 for `non-white' subjects and 1 for `white' subjects.
		
		\item \textit{COMPAS} \cite{larson2016we}: 
		The Compas Propublica Risk Assessment dataset contains 6,172 subjects to predict recidivism (`HighScore' or `LowScore') with 6 variables related to criminal history and demographic information. We use racial characteristics as a sensitive variable.
		
	\end{itemize}
	We transform all categorical variables to dummy variables using one-hot encoding, and standardize to get zero mean and 1 standard deviation for each variable.
	Some variables having serious multicollinearity have been removed in order to obtain stable estimation results.
	The performances of the unconstrained linear logistic model are summarized in Table \ref{sample-table}.
	\begin{table}[t]
		\caption{Performances of the unconstrained linear logistic model on the test dataset. Except for \textit{Adult}, average metrics are described.}
		\label{sample-table}
		\begin{center}
			\begin{footnotesize}
				\begin{sc}
					\begin{tabular}{llcccc}
						\toprule
						Model  & Dataset & Acc & DI & EOp & DM\\ 
						\midrule
						Linear & \textit{Adult}    & 0.852 & 0.172 & 0.070 & 0.117 \\
						& \textit{Bank}      & 0.908 & 0.195 & 0.099 & 0.176 \\
						& \textit{LSAC}      & 0.823 & 0.120 & 0.041 & 0.090 \\
						& \textit{COMPAS}    & 0.757 & 0.164 & 0.074 & 0.020 \\
						\midrule
						DNN    & \textit{Adult}   & 0.853 & 0.170 & 0.076 & 0.105 \\
						& \textit{Bank}     & 0.904 & 0.236 & 0.082 & 0.174 \\
						& \textit{LSAC}     & 0.856 & 0.131 & 0.038 & 0.071 \\
						& \textit{COMPAS}   & 0.757 & 0.162 & 0.075 & 0.024 \\
						\bottomrule
					\end{tabular}
				\end{sc}
			\end{footnotesize}
		\end{center}
	\end{table}
	
	\subsection{Implementation details}
	For numerical stability, we use the ridge penalty for DNN parameters with the regularization parameter $10^{-6}$.
	All experiments are conducted on a GPU server with NVIDIA TITAN Xp GPUs.
	Also, for each method, we consider $\text{lr} \in \{0.01, 0.1, 1\}$ and $\text{epoch} \in \{10000, 20000 \}$, then we choose the best learning rate and epoch. 
	In addition, we did not use a mini-batch for the gradient descent approach, i.e., we set the batch size to the sample size.
	For each BGF constraint, we choose the corresponding regularization parameter so that the value of the BGF constraint (e.g., DI, EOp, MSP) reaches a certain level among the following candidate parameters set:
	$$\lambda \in \{ 0, 0.05, 0.1, 0.35, 0.45, 0.6, 0.75, 1, 2, 5\}. $$
	The hyper-parameters in the doubly-fair algorithm are set to minimize the dWGF (or WGF) value while the BGF level remains similar to that of the BGF classifier, among the following candidate parameters sets:
	\begin{equation*}
		\begin{split}
			\lambda &\in \{ 0, 0.05, 0.1, 0.35, 0.45, 0.6, 0.75, 1, 2, 5\} \\
			\eta &\in \{ 0, 0.1, 0.5, 1, 3, 5 \}.
		\end{split}
	\end{equation*}
	
	For the WGF score function, we adopt the surrogated version of Kendall's $\tau$ as the WGF constraint.
	However, the surrogated Kendall's $\tau$ requires huge computation since it should process
	all pairs of the training data.
	To save computing time for calculating the surrogated Kendall's $\tau$, we
	use 50,000 pairs of samples randomly selected from the training data for each sensitive group.

\end{document}